\documentclass[10pt,twocolumn,letterpaper]{article}

\usepackage{iccv}
\usepackage{times}
\usepackage{epsfig}
\usepackage{graphicx}
\usepackage{amsmath}
\usepackage{amssymb}

\usepackage{booktabs}
\usepackage{multirow}
\usepackage{bm}
\usepackage{subcaption}
\usepackage{makecell}
\usepackage{algorithm,algorithmic}
\usepackage{enumitem}

\newcommand{\name}{CMOT}
\newcommand{\UCF}{UCF-101}
\newcommand{\HMDB}{HMDB-51}
\newcommand{\SMSM}{SMSM-V2}

\newcommand{\red}[1]{\textcolor{red}{#1}}
\newcommand{\blue}[1]{\textcolor{blue}{#1}}

\newcommand{\tabincell}[2]{\begin{tabular}{@{}#1@{}}#2\end{tabular}}

\newcommand{\figref}[1]{Figure~\ref{#1}}
\newcommand{\tabref}[1]{Table~\ref{#1}}
\newcommand{\secref}[1]{Section~\ref{#1}}
\newcommand{\algref}[1]{Algorithm~\ref{#1}}
\newcommand{\equref}[1]{Equation~(\ref{#1})}

\newtheorem{definition}{Definition}
\newtheorem{theorem}{Theorem}

\usepackage[pagebackref=true,breaklinks=true,letterpaper=true,colorlinks,bookmarks=false]{hyperref}

\iccvfinalcopy 

\def\iccvPaperID{3279} 

\ificcvfinal\fi

\begin{document}

\title{Few-Shot Action Recognition with Compromised Metric via Optimal Transport}

\author{
Su Lu~~~~~~~~~~~~Han-Jia Ye~~~~~~~~~~~~De-Chuan Zhan\\
State Key Laboratory for Novel Software Technology\\
Nanjing University, Nanjing, China\\
{\tt\small \{lus,yehj\}@lamda.nju.edu.cn, zhandc@nju.edu.cn}
}

\maketitle
\ificcvfinal\thispagestyle{empty}\fi

\begin{abstract}
Although vital to computer vision systems, few-shot action recognition is still not mature despite the wide research of few-shot image classification.
Popular few-shot learning algorithms extract a transferable embedding from seen classes and reuse it on unseen classes by constructing a metric-based classifier.
One main obstacle to applying these algorithms in action recognition is the complex structure of videos.
Some existing solutions sample frames from a video and aggregate their embeddings to form a video-level representation, neglecting important temporal relations. Others perform an explicit sequence matching between two videos and define their distance as matching cost, imposing too strong restrictions on sequence ordering.
In this paper, we propose \textbf{C}ompromised \textbf{M}etric via \textbf{O}ptimal \textbf{T}ransport~({\name}) to combine the advantages of these two solutions. {\name} simultaneously considers semantic and temporal information in videos under Optimal Transport framework, and is discriminative for both content-sensitive and ordering-sensitive tasks.
In detail, given two videos, we sample segments from them and cast the calculation of their distance as an optimal transport problem between two segment sequences. To preserve the inherent temporal ordering information, we additionally amend the ground cost matrix by penalizing it with the positional distance between a pair of segments.
Empirical results on benchmark datasets demonstrate the superiority of {\name}.
\end{abstract}

\section{Introduction} \label{Section:introduction}
The ability to learn with limited data is important, especially when data collection is difficult. In computer vision systems, it may be impossible to obtain abundant images of rare targets, restricting the application of image classification~\cite{FEAT}, object detection~\cite{RareOD1,RareOD2} and image segmentation~\cite{RareIS}. Action recognition means recognizing a human action from a video containing complete action execution~\cite{ARSurvey2}, which is another non-negligible technology in computer vision. It extends over many real-world applications, from human-computer interaction to video surveillance and information retrieval~\cite{ARSurvey1}. Reducing the demand for training instances in action recognition is critical owing to the high expense of gathering annotated videos.

Few-shot learning aims to endow a learner with the ability to recognize new classes from a small number of labeled examples. Directly training a model with limited instances easily falls into the dilemma of over-fitting. Thus, we often assume that another large related dataset containing seen classes is available, and the target is to classify unseen classes given a few annotated instances. Among few-shot learning algorithms, metric-based algorithms have achieved promising results~\cite{MatchNet,ProtoNet,TADAM,FEAT}. These methods try to learn a generalizable embedding function from seen classes and reuse it to measure instance distances or similarities on unseen classes. We can build a metric-based classifier~({\em e.g.}, nearest neighbour classifier) to make predictions with assistance of the learned embedding function. Although successful in few-shot image classification, the aforementioned algorithms are not specially designed for video data, making it hard to reuse them in few-shot action recognition.

\begin{figure}[!t]
	\centering
	\includegraphics[width=\columnwidth]{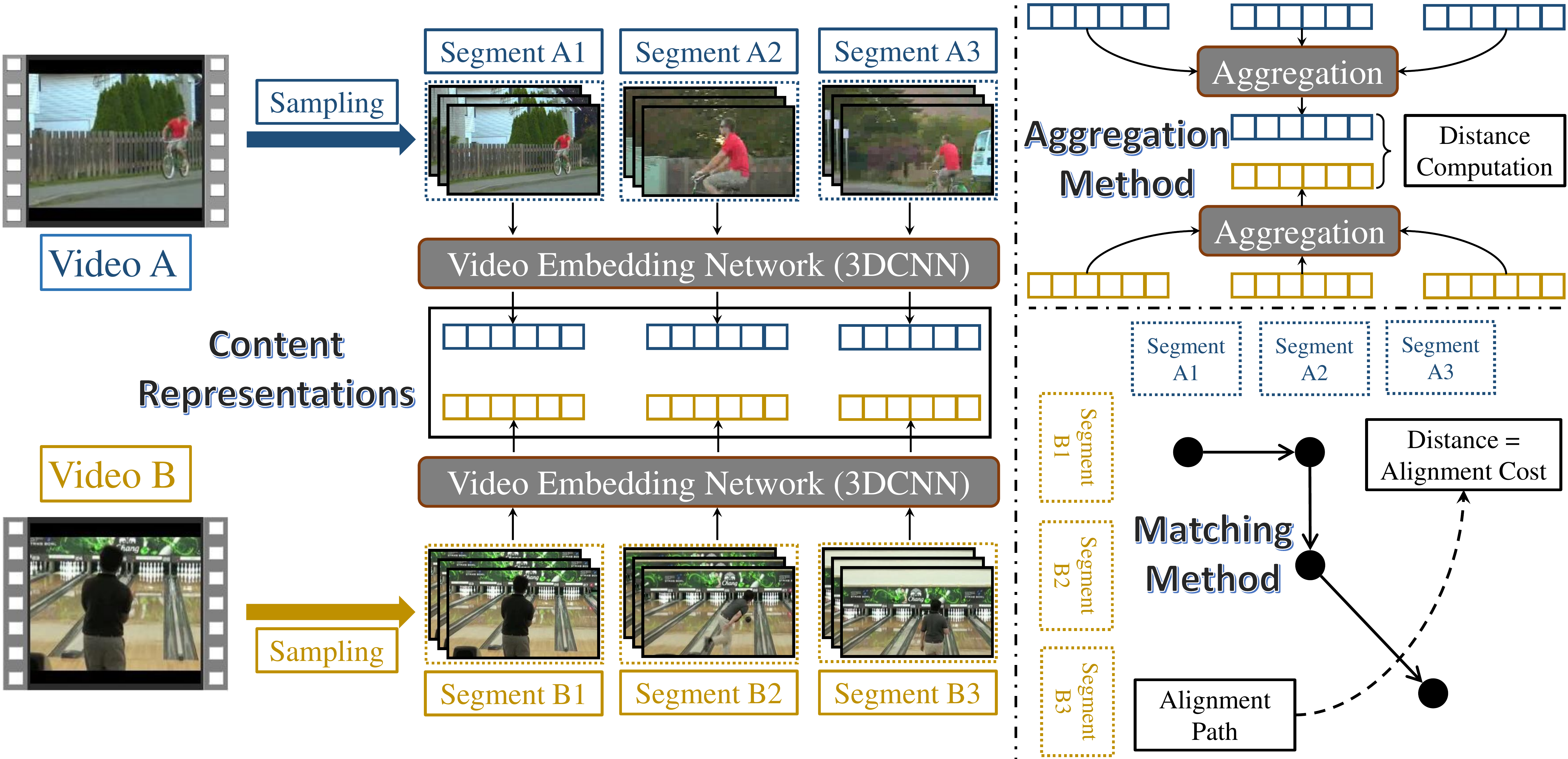}
	\caption{Illustration of aggregation-based methods and matching-based method. Aggregation-based methods use aggregation function like average pooling to obtain a video-level representation, neglecting temporal information. Matching-based methods find a strictly progressive alignment between segment sequences.}
	\label{Figure:aggregaion_matching}
\end{figure}

Some existing methods are direct extensions from few-shot image classification algorithms~\cite{TARN,TAEN,TAV}. They first obtain semantic representations of video frames~(or video segments) by a CNN~(or 3DCNN), and then aggregate the sequence of vectors into a single video embedding. The simplest aggregation function is average pooling~\cite{MB}, which naively sums up the vectors in the sequence. Attention is also widely used when aggregating segment representations~\cite{PASTN,ARN}. By considering the similarities between segments, attention can output a weighted sum of segment features. However, these \textbf{aggregation-based} methods ignore the long-term temporal relations between segments, resulting in a sub-optimal measurement of video distances. Consider two different actions, {\em i.e.}, moving an object from left to right and moving an object from right to left. Since the aggregation process neglects the ordering of segments, the distance between these two actions will be small, which is harmful to the performance of classifier. Figure~\ref{Figure:aggregaion_matching}~(upper right part) is an illustration of \textbf{aggregation-based} methods.

To overcome this issue, researchers have proposed to explicitly match two videos by aligning their segment sequences without obtaining video-level representations. In OTAM~\cite{OTAM}, the authors use Dynamic Time Warping~(DTW) to find the best alignment between two sequences and define the distance between them as the alignment cost. As shown in Figure~\ref{Figure:aggregaion_matching}~(bottom right part), this kind of \textbf{matching-based} methods ensures that the long-term ordering information is considered during the computation of video distances. For example, if we inverse the frame sequence of a video to generate a new video, the matching cost between them will be large and OTAM tends to predict them as different classes. This characteristic helps to differ actions that are sensitive to long-term ordering~({\em e.g.}, aforementioned left2right and right2left). However, for actions that are irrelevant to temporal ordering but sensitive to content, OTAM may work badly owing to DTW's strong restriction on matching rules.

From the above introduction about two kinds of methods, we can see that they are complementary. \textbf{Aggregation-based} methods focus on semantic information while \textbf{matching-based} methods emphasize temporal ordering. Actually, both semantic and temporal information should be considered in few-shot action recognition. In this paper, we propose \textbf{C}ompromised \textbf{M}etric via \textbf{O}ptimal \textbf{T}ransport~({\name}), which simultaneously compares two videos' content difference and ordering difference to give a compromised measurement under Optimal Transport~(OT) framework~\cite{OT1,OT2}. {\name} balances semantic and temporal information in videos, achieving competitive performance in both content-sensitive tasks and ordering-sensitive tasks.

In {\name}, we first sample several segments from a video, and get their embeddings by 3DCNN~\cite{C3D,I3D} to form a sequence of content representations. Given two videos, we can compute the semantic cost matrix between their content representations. To preserve the inherent temporal ordering information, we additionally amend the semantic cost matrix by penalizing it with the positional distance between a pair of segments. To be specific, the cost between two segments at relatively far positions~({\em e.g.}, the first segment of a video and the last segment of another one) will be enlarged. Performing OT on such a fused cost matrix induces an optimal transportation plan between two videos based on both semantic and temporal information. We then take the transportation cost as two videos' distance, and build a metric-based classifier to optimize the 3DCNN.

We demonstrate the superiority of balancing semantic and temporal information in a synthetic experiment. {\name} also achieves competitive performance on several benchmark datasets, {\em e.g.}, {\UCF}~\cite{UCF-101}, {\HMDB}~\cite{HMDB-51}, and {\SMSM}~\cite{SMSM-V2}. Our contributions are threefold:
\begin{itemize}[leftmargin=1mm]
\item We jointly consider semantic and temporal information.
\item We formulate the distance between two videos as transportation cost, and encapsulate the OT algorithm into metric-based few-shot learning framework.
\item We empirically demonstrate the effectiveness of {\name} on synthetic experiment and action recognition datasets.
\end{itemize}

\section{Related Work}\label{Section:related_work}
\paragraph{Meta-Learning.} Meta-Learning~\cite{Learning2Learn} aims at extracting task-level experience from seen data, while generalizing the learned meta-knowledge to unseen tasks efficiently. It acts as one main tool for few-shot learning. Meta-learning algorithms can be classified into three categories based on what kind of meta-knowledge they catch. Memory-based methods~\cite{NTM,MANN,MNM} use external memory to facilitate task learning. Optimization-based methods~\cite{MAML,Meta-SGD,Meta-LSTM,MUMOMAML,LEO} try out some optimization strategies and accelerate the training process of unseen tasks. Our proposed method falls into the remaining category, metric-based algorithms~\cite{MatchNet,ProtoNet,MetaOptNet,Siamese,FEAT,TADAM,CTM,DSN,CAN}, which offer a shared embedding for instances from new tasks. Metric-based methods are suitable for video problems because they are more efficient than others. In our proposed {\name}, two videos' distance is defined as the optimal transportation cost based on a fused cost matrix. We further encapsulate the OT algorithm into the metric-based meta-learning framework to make the whole model differentiable~\cite{DeepEMD}.
\paragraph{Few-Shot Action Recognition.}
Following the idea of memory-based meta-learning, CMN~\cite{CMN} learns several key-value pairs to represent a video in a large space. Apart from CMN, most of the existing few-shot action recognition algorithms follow the idea of metric-based meta-learning. They extract a generalizable video metric on previous seen tasks, and transfer it to unseen new tasks. This line of methods can be classified into two categories, {\em i.e.}, aggregation-based methods and matching-based methods. Both of these two kinds of methods first sample frames or segments from a video and get their content representations, but aggregation-based methods generate a video-level representation for distance calculation by pooling~\cite{MB,ProtoGAN,3DFSV,PAL,TAEN,ITA}, adaptive fusion~\cite{AMeFu,TAV}, dynamic images~\cite{FAN}, or attention~\cite{TARN,ARN,PASTN} while matching-based methods explicitly align two sequences and define distance between videos as their matching cost by DTW~\cite{OTAM} or other matching strategies. Aggregation-based methods focus on semantic contents while matching-based methods emphasize temporal ordering, and our proposed {\name} simultaneously considers two aspects to form a compromised metric.

\section{Preliminary}\label{Section:preliminary}
In this section, we will describe two basic components of our proposed method. Firstly, we will introduce the problem setting of few-shot learning, and present a simple solution by transferring the embedding function~\cite{ProtoNet}. Secondly, we give a concise review of optimal transport~\cite{OT1,OT2}.
\subsection{Few-Shot Learning}
Few-shot learning means learning from limited examples. In action recognition scenario, an $N$-way $K$-shot task is composed of $N$ classes and $K$ training videos per class. Another testing set sampled from the same $N$ classes is provided to evaluate the classifier. In few-shot learning literature, the small training set of each task is referred as {\em support} set $\mathcal{S}=\{(\mathbf{x}_i,y_i)\}_{i=1}^{NK}$ and the testing set is called {\em query} set $\mathcal{Q}=\{(\mathbf{x}_j,y_j)\}_{j=1}^{NQ}$. That is, a task $\mathcal{T}$ is defined as $\mathcal{T}=(\mathcal{S},\mathcal{Q})$. We assume that each video $\mathbf{x}$ is composed of $M$ segments $\mathbf{x}=[\mathbf{x}^{1},\cdots,\mathbf{x}^{m},\cdots,\mathbf{x}^{M}]$, and each segment $\mathbf{x}^{m}$ contains a certain number of consecutive frames.

\begin{figure}[!t]
	\centering
	\includegraphics[width=\columnwidth]{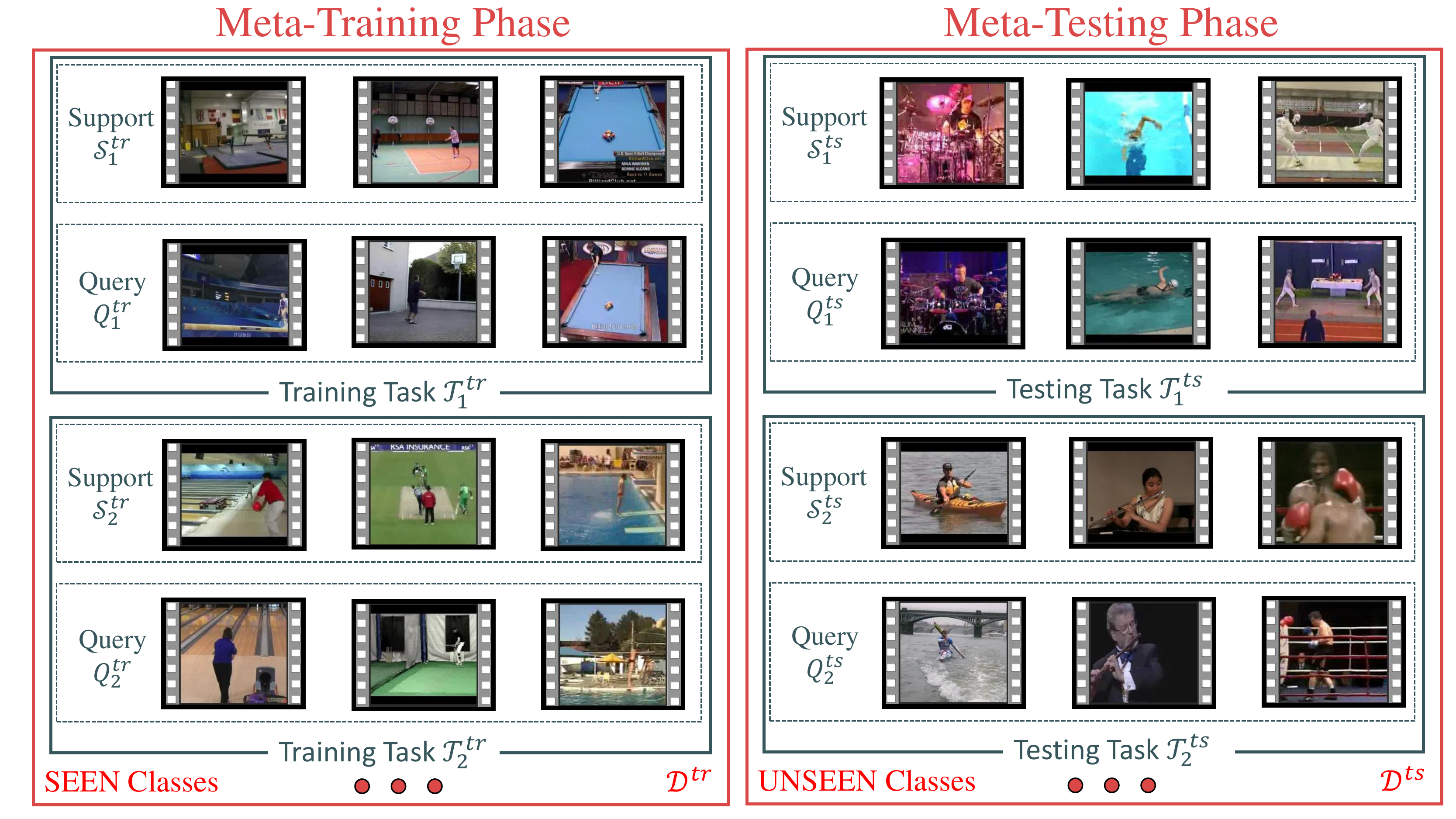}
	\caption{An illustration of episodic training protocol. During {\em meta-training} phase, $N$-way $K$-shot action recognition tasks are sampled from $\mathcal{D}^{tr}$ and used to optimize the meta-model. During {\em meta-testing} phase, we sample $N$-way $K$-shot tasks with non-overlapping classes from $\mathcal{D}^{ts}$. The target is to obtain high accuracy on unseen {\em meta-testing} tasks.}
	\label{Figure:episodic}
\end{figure}

Researchers often utilize meta-learning to tackle few-shot action recognition problems. A key idea in meta-learning is to mimic {\em meta-testing} process in {\em meta-training} phase. Since the learned meta-model is intended for $N$-way $K$-shot classification tasks, we sample episodic $N$-way $K$-shot tasks from {\em meta-training} set $\mathcal{D}^{tr}$~(composed of SEEN classes) to optimize our model. The main target is to extract knowledge from sampled tasks and reuse it when a new task comes. In {\em meta-testing} phase, $N$-way $K$-shot tasks are sampled from a {\em meta-testing} set $\mathcal{D}^{ts}$~(composed of UNSEEN classes) to evaluate model performance. \figref{Figure:episodic} gives an illustration of this protocol.

A simple solution is to meta-learn an embedding function $\phi$, which maps an input video segment $\mathbf{x}^{m}$ to a vectorial representation $\phi(\mathbf{x}^{m})$. With a bit abuse of notations, we denote by $\phi(\mathbf{x})$ the sequence of embedded segments $[\phi(\mathbf{x}^{1}),\cdots,\phi(\mathbf{x}^{m}),\cdots,\phi(\mathbf{x}^{M})]$. In a {\em meta-training} task $\mathcal{T}^{tr}$, the label of a {\em query} video $\mathbf{x}_j$ could be determined by its mean distance to videos belonging to each class in the {\em support} set $\mathcal{S}^{tr}$ as shown in the following two equations. $\mathrm{dis}(\cdot,\cdot)$ is some function measuring the distance between two transformed segment sequences.

\begin{equation}
p(\hat{y}_j=n|\mathbf{x}_j) = \frac{\exp\left\{-\overline{\mathrm{DIS}}(\phi(\mathbf{x}_j);n)\right\}}{\sum_{n^\prime=1}^{N}\exp\left\{-\overline{\mathrm{DIS}}(\phi(\mathbf{x}_j);n^\prime)\right\}}
\label{Equation:protonet_inference}
\end{equation}

\begin{equation}
\overline{\mathrm{DIS}}(\phi(\mathbf{x}_j);n)=\frac{1}{K}\sum_{(\mathbf{x}_i,y_j)\in\mathcal{S}^{tr}\wedge y_i=n}\mathrm{dis}(\phi(\mathbf{x}_i),\phi(\mathbf{x}_j))
\label{Equation:protonet_center}
\end{equation}
Cross-entropy loss is optimized on all sampled tasks.

\begin{equation}
\min_{\phi} \sum_{\mathcal{T}^{tr} \sim \mathcal{D}^{tr}} \sum_{(\mathbf{x}_j,y_j)\in\mathcal{Q}^{tr}} -\log p(\hat{y}_j=y_j|\mathbf{x}_j)
\label{Equation:protonet_objective}
\end{equation}
After {\em meta-training} phase, we apply the learned embedding function $\phi$ to $N$-way $K$-shot tasks $\mathcal{T}^{ts}$ sampled from $\mathcal{D}^{ts}$.

Obviously, the distance function $\mathrm{dis}(\cdot,\cdot)$ will influence the model performance to a great extent. Aggregation-based methods~\cite{ARN} define $\mathrm{dis}(\cdot,\cdot)$ as the Euclidean distance between aggregated video representations, {\em i.e.}, $\mathrm{dis}^{\mathrm{AGG}}(\phi(\mathbf{x}_{i}),\phi(\mathbf{x}_{j}))=\|\mathrm{AGG}(\phi(\mathbf{x}_{i}))-\mathrm{AGG}(\phi(\mathbf{x}_{j}))\|_2$ where $\mathrm{AGG}$ is an aggregation function like average pooling. Important temporal relations in $\phi(\mathbf{x}_i)$ and $\phi(\mathbf{x}_j)$ are neglected. Matching-based methods~\cite{OTAM} define $\mathrm{dis}(\cdot,\cdot)$ as the matching cost between two sequences $\phi(\mathbf{x}_i)$ and $\phi(\mathbf{x}_j)$, {\em i.e.}, $\mathrm{dis}^{\mathrm{MAT}}(\phi(\mathbf{x}_{i}),\phi(\mathbf{x}_{j}))=\mathrm{MAT}(\phi(\mathbf{x}_i),\phi(\mathbf{x}_j))$. By performing sequence alignment algorithm such as DTW, matching-based methods find a strictly progressive route from the first segment to the last segment, imposing too strong restrictions on the ordering of videos. In {\name}, we specially design a distance function that combines two methods under a unifying OT framework.
\subsection{Optimal Transport}
The theory of optimal transport defines a geometry to compare measures supported on metric probability spaces~\cite{OT3}. It finds a best matching between two distributions to minimize the transportation cost given a cost matrix, based on which we can define a distance between these two distributions. OT is used in many applications, {\em e.g.}, semantic correspondence analysis~\cite{OT_Semantic}, unsupervised domain adaptation~\cite{OT_UDA}, and label distribution learning~\cite{OT_Label}. In this paper, we focus on discrete optimal transport.

\begin{definition}[Transportation Plan]
For two discrete distributions $\bm{\mu},\bm{\nu}\in\Delta_{d}$ where $\Delta_{d}=\{\bm{\alpha}\in\mathbb{R}_{+}^{d}|\bm{\alpha}^\top\mathbf{1}_{d}=1\}$\footnote{$\mathbb{R}_{+}$ is the set of non-negative real numbers.} is the set of $d$-dimensional probability simplexes, let $\Pi(\bm{\mu},\bm{\nu})$ be the set of transport polytopes, which contains all legal transportation plans from $\bm{\mu}$ to $\bm{\nu}$.
\begin{equation}
\Pi(\bm{\mu},\bm{\nu})=\{\mathbf{T}\in\mathbb{R}_{+}^{d\times d}|\mathbf{T}\mathbf{1}_{d}=\bm{\mu},\mathbf{T}^\top\mathbf{1}_{d}=\bm{\nu}\}
\label{Equation:ot_transport_polytope}
\end{equation}
\end{definition}

\begin{definition}[Optimal Transport]
Given a cost matrix $\mathbf{C}\in\mathbb{R}^{d\times d}$, the total cost of mapping from $\bm{\mu}$ to $\bm{\nu}$ using transportation plan $\mathbf{T}$ can be quantified as $\langle\mathbf{T},\mathbf{C}\rangle$. The distance between $\bm{\mu}$ and $\bm{\nu}$ is defined as an OT problem.
\begin{equation}
\eta_{\mathbf{C}}\triangleq\min_{\mathbf{T}\in\Pi(\bm{\mu},\bm{\nu})}\langle\mathbf{T},\mathbf{C}\rangle
\label{Equation:ot_distance}
\end{equation}
\end{definition}

The optimization problem defined in \equref{Equation:ot_distance} is a classic optimal transport formulation. However, solving such a problem is computationally expensive. According to~\cite{OT2}, adding an entropy regularization term to the transport object gives rise to a smoothed version of the underlying problem. This makes the objective function strictly convex, and we have efficient algorithms to solve it.

\begin{definition}[Sinkhorn Distance]
Define the entropy of $\mathbf{T}$ as $\mathcal{H}(\mathbf{T})=-\sum_{i=1}^{d}\sum_{j=1}^{d}\mathbf{T}_{ij}\log(\mathbf{T}_{ij})$. By solving a smoothed version of optimal transport problem, we have the Sinkhorn distance between $\bm{\mu}$ and $\bm{\nu}$.
\begin{equation}
\eta_{\mathbf{C}}^{\lambda}\triangleq\min_{\mathbf{T}\in\Pi(\bm{\mu},\bm{\nu})}\langle\mathbf{T},\mathbf{C}\rangle-\frac{1}{\lambda}\mathcal{H}(\mathbf{T})
\label{Equation:ot_sinkhorn}
\end{equation}
\end{definition}

\begin{theorem}[Sinkhorn Algorithm~\cite{OT2}]
Define matrix $\mathbf{G}=\exp(-\lambda\mathbf{C})$ as the element-wise exponential of $\lambda\mathbf{C}$. The optimizer $\mathbf{T}^\star$ of problem~\equref{Equation:ot_sinkhorn} can be represented as the following form, where vector $\mathbf{u}$ and $\mathbf{v}$ can be obtained by an iterative algorithm $\mathbf{u}\leftarrow\bm{\mu}\oslash(\mathbf{G}\mathbf{v})$,$\mathbf{v}\leftarrow\bm{\nu}\oslash(\mathbf{G}^\top\mathbf{u})$.\footnote{$\oslash$ mean element-wise division.} Refer to the supplement for proof and more details.
\begin{equation}
\mathbf{T}^\star=\mathrm{diag}(\mathbf{u})\mathbf{G}\mathrm{diag}(\mathbf{v})
\label{Equation:ot_solution}
\end{equation}
\end{theorem}

\section{Main Approach}\label{Section:main_approach}
In this section, we will describe our proposed {\name}. Firstly, we will introduce segment descriptor for extracting content representations, and then show how to use optimal transport to measure the semantic gap between two videos. Secondly, we add a soft restriction on video ordering by penalizing the ground cost matrix with positional distance between a pair of segments. We show that semantic and temporal information can be jointly considered in a unifying framework. Finally we formulate the entire {\name}.

\begin{figure*}[!t]
	\centering
	\includegraphics[width=\textwidth]{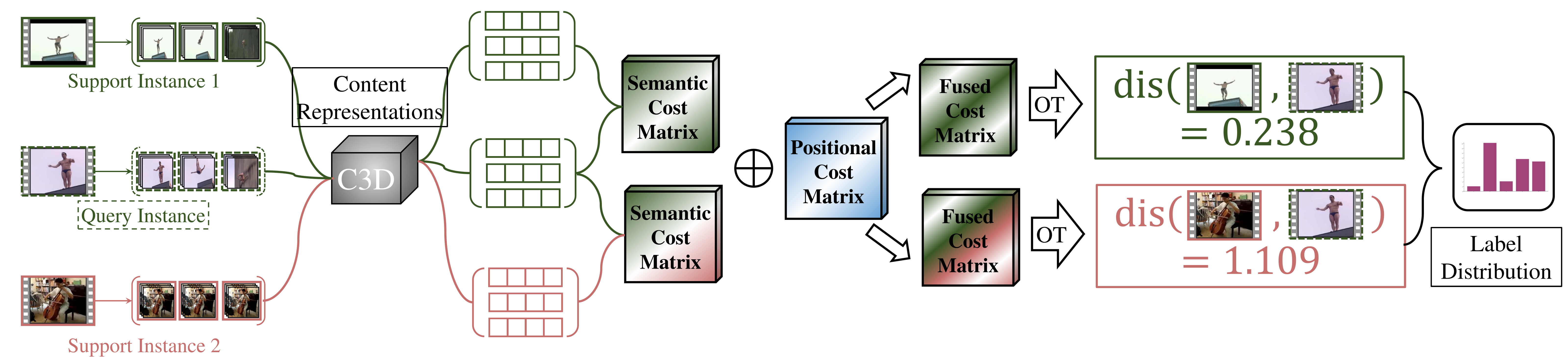}
	\caption{Framework of {\name}. In this figure, we show a $2$-way $1$-shot task. We firstly sample segments from each video, and then get their content representations by C3D. After that, we compute the fused cost matrix which jointly considers semantic and temporal information. Optimal transport is used to find a matching between segment sequences, based on which we can get {\em query} instance's distances to both {\em support} instances.}
	\label{Figure:dom_framework}
\end{figure*}

\subsection{Semantic Matching}
In this part, we will describe how to use optimal transport to match the contents of two videos. After extracting segment embeddings by 3DCNN~\cite{C3D,I3D}, we formulate the semantic distance between two videos as the optimal transportation cost between their segment sequences.
\paragraph{Segment Descriptor.} As we know, a video describing an action may be split into several segments, each of which corresponds to a sub-action. Studying the short-term sub-actions benefits to the recognition of human motions. In {\name}, we assume all the videos at hand have similar lengths, and each video $\mathbf{x}$ can be represented by a series of ordered segments, {\em i.e.}, $\mathbf{x}=[\mathbf{x}^{1},\cdots,\mathbf{x}^{m},\cdots,\mathbf{x}^{M}]$.\footnote{For simplicity we assume all the videos have $M$ segments, but our method can handle videos that have different numbers of segments without any modification.} With assistance of a 3DCNN $\phi$, we are able to embed a short segment $\mathbf{x}^{m}$ to a $D$-dimensional vector $\phi(\mathbf{x}^{m})$. Thus, the entire video is represented as $[\phi(\mathbf{x}^{1}),\cdots,\phi(\mathbf{x}^{m}),\cdots,\phi(\mathbf{x}^{M})]$. With a bit abuse of notations, we denote the transformed segment sequence of length $M$ by $\phi(\mathbf{x})$.
\paragraph{How to transport between videos?} Following the thoughts in~ \secref{Section:preliminary}, now we need to design a distance function $\mathrm{dis}(\cdot,\cdot)$ to measure the semantic gap between two segment sequences. First of all, we introduce a probabilistic viewpoint of instance generation, which is widely adopted by machine learning society. For example, in NLP, a sentence can be defined as a probability distribution over words~\cite{S-WMD}. Here we define each video as a distribution over segments, and then use Sinkhorn distance~\equref{Equation:ot_sinkhorn} to measure the dissimilarity between two videos. The cost matrix can be obtained by computing Euclidean distances between segment features. To summarize, the semantic distance between two videos $\mathbf{x}_1$ and $\mathbf{x}_2$ are formalized as~\equref{Equation:segment_descriptor_distance} and~ \equref{Equation:segment_descriptor_cost}, where $\bm{\mu}_1$ and $\bm{\mu}_2$ are segment distributions of $\mathbf{x}_1$ and $\mathbf{x}_2$ respectively. Without any prior knowledge, we can set $\bm{\mu}_1$ and $\bm{\mu}_2$ as $M$-dimensional uniform distribution. If we know that some segment are more important in a video, we can increase their weights by  conveniently tuning the segment distributions.
\begin{equation}
\mathrm{dis}^{\mathrm{SE}}(\mathbf{x}_1,\mathbf{x}_2)=\min_{\mathbf{T}\in\Pi(\bm{\mu}_{1},\bm{\mu}_{2})}\langle\mathbf{T},\mathbf{C}^{\mathrm{SE}}\rangle-\frac{1}{\lambda}\mathcal{H}(\mathbf{T})
\label{Equation:segment_descriptor_distance}
\end{equation}
\begin{equation}
\mathbf{C}^{\mathrm{SE}}_{pq}=\|\phi(\mathbf{x}_1^{p})-\phi(\mathbf{x}_2^{q})\|_2,\forall p,q\in[M]
\label{Equation:segment_descriptor_cost}
\end{equation}

\subsection{Temporal Modelling}
As indicated in \secref{Section:preliminary}, a key point in few-shot action recognition is the design of distance metric. Like aggregation-based methods~\cite{TAV}, the OT-based method described in the previous subsection matches two videos' contents, neglecting important temporal relations in videos. In this part, we further model temporal relations by optimal transport, imposing a soft regulation on video orderings.
\paragraph{Positional Cost Matrix.} The purpose of considering long-term relations is to ensure that a segment in $\mathbf{x}_1$ is mapped to a segment in $\mathbf{x}_2$ at near positions. This helps to differ actions that are sensitive to orderings. For example, consider two actions: moving an object from left to right~(left2right) and moving an object from right to left~(right2left). If we ignore the ordering of sub-actions and only focus on video contents, these two actions will not be separated easily. In OTAM~\cite{OTAM}, the authors proposed to use DTW to align two videos. DTW is a dynamic programming algorithm and is strictly progressive in time dimension. That is, for two segments $\mathbf{x}_1^1$ and $\mathbf{x}_1^2$ in video $\mathbf{x}_1$, if $\mathbf{x}_1^1$ is located before $\mathbf{x}_1^2$, it must be mapped to a segment in video $\mathbf{x}_2$ before the target segment of $\mathbf{x}_1^2$. This characteristic imposes too strong restrictions on the ordering of videos, and is harmful to the robustness of the learned distance metric. In this paper, we define a positional cost matrix $\mathbf{C}^{\mathrm{PO}}$ as \equref{Equation:positional_cost_matrix}, whose value increases with the relative positional distance $\left(\frac{p}{M}-\frac{q}{M}\right)^2$. Here $\sigma$ is a hyper-parameter.
\begin{equation}
\mathbf{C}^{\mathrm{PO}}_{pq}=\exp\left\{-\frac{1}{\sigma^2}\frac{1}{\left(\frac{p}{M}-\frac{q}{M}\right)^2+1}\right\},\forall p,q\in[M]
\label{Equation:positional_cost_matrix}
\end{equation}
Similar to~\equref{Equation:segment_descriptor_distance}, we can further define two videos' positional distance as~\equref{Equation:positional_distance}.
\begin{equation}
\mathrm{dis}^{\mathrm{PO}}(\mathbf{x}_1,\mathbf{x}_2)=\min_{\mathbf{T}\in\Pi(\bm{\mu}_{1},\bm{\mu}_{2})}\langle\mathbf{T},\mathbf{C}^{\mathrm{PO}}\rangle-\frac{1}{\lambda}\mathcal{H}(\mathbf{T})
\label{Equation:positional_distance}
\end{equation}
Obviously, $\mathbf{C}^{\mathrm{PO}}$ imposes a soft regularization on the ordering of videos by assigning a larger transportation cost between distant segments.

\subsection{{\name} Framework}
In this subsection, we show that semantic matching and temporal matching can be unified under a single OT problem, and give the formal expression of {\name}.
\paragraph{Compromised Distance Metric.} By simultaneously considering semantic contents and temporal orderings, we can easily come up with a compromised distance function, {\em i.e.}, the weighted sum of $\mathrm{dis}^{\mathrm{SE}}(\cdot,\cdot)$ and $\mathrm{dis}^{\mathrm{PO}}(\cdot,\cdot)$:
\begin{equation}
\mathrm{dis}(\mathbf{x}_1,\mathbf{x}_2)=\mathrm{dis}^{\mathrm{SE}}(\mathbf{x}_1,\mathbf{x}_2)+\alpha\mathrm{dis}^{\mathrm{PO}}(\mathbf{x}_1,\mathbf{x}_2)
\label{Equation:compromised_distance}
\end{equation}
In~\equref{Equation:compromised_distance}, $\alpha$ is a hyper-parameter balancing the importance of two parts. By the linearity of matrix inner-production, we can rewrite the distance in~\equref{Equation:compromised_distance} as the following OT problem.
\begin{equation}
\begin{aligned}
&\mathrm{dis}(\mathbf{x}_1,\mathbf{x}_2)&&=\min_{\mathbf{T}\in\Pi(\bm{\mu}_{1},\bm{\mu}_{2})}\langle\mathbf{T},\mathbf{C}\rangle-\frac{1}{\lambda}\mathcal{H}(\mathbf{T})\\
&\mathrm{s.t.}&&\mathbf{C}=\mathbf{C}^{\mathrm{SE}}+\alpha\mathbf{C}^{\mathrm{PO}}
\end{aligned}
\label{Equation:compromised_distance_ot}
\end{equation}

\paragraph{Objective.} We now summarize the formal expression of {\name}. For an $N$-way $K$-shot task $\mathcal{T}^{tr}=(\mathcal{S}^{tr},\mathcal{Q}^{tr})$ sampled from {\em meta-training} set $\mathcal{D}^{tr}$, the label of {\em query} instance $\mathbf{x}_j$ is predicted as~\equref{Equation:cmot_inference} and~\equref{Equation:cmot_center}.
\begin{equation}
p(\hat{y}_j=n|\mathbf{x}_j) = \frac{\exp\left\{-\overline{\mathrm{DIS}}(\mathbf{x}_j;n)\right\}}{\sum_{n^\prime=1}^{N}\exp\left\{-\overline{\mathrm{DIS}}(\mathbf{x}_j;n^\prime)\right\}}
\label{Equation:cmot_inference}
\end{equation}
\begin{equation}
\overline{\mathrm{DIS}}(\mathbf{x}_j;n)=\frac{1}{K}\sum_{(\mathbf{x}_i,y_j)\in\mathcal{S}^{tr}\wedge y_i=n}\mathrm{dis}(\mathbf{x}_i,\mathbf{x}_j)
\label{Equation:cmot_center}
\end{equation}
Here $\mathrm{dis}(\mathbf{x}_i,\mathbf{x}_j)$ is defined as~\equref{Equation:compromised_distance}, {\em i.e.}, Sinkhorn distance based on fused cost matrix. The main objective of {\name} is shown as~\equref{Equation:cmot_objective}.
\begin{equation}
\begin{aligned}
&\min_{\phi}&&\sum_{\mathcal{T}^{tr} \sim \mathcal{D}^{tr}} \sum_{(\mathbf{x}_j,y_j)\in\mathcal{Q}^{tr}} -\log p(\hat{y}_j=y_j|\mathbf{x}_j) \\
&\mathrm{s.t.}&&\mathrm{dis}(\mathbf{x}_i,\mathbf{x}_j)=\min_{\mathbf{T}\in\Pi(\bm{\mu}_{i},\bm{\mu}_{j})}\langle\mathbf{T},\mathbf{C}\rangle-\frac{1}{\lambda}\mathcal{H}(\mathbf{T})\\
&&&\textbf{C}=\textbf{C}^{\mathrm{SE}}+\alpha\mathbf{C}^{\mathrm{PO}}
\end{aligned}
\label{Equation:cmot_objective}
\end{equation}
\figref{Figure:dom_framework} shows the whole framework of {\name}.

\paragraph{Solution to Bi-Level Optimization.} Obviously, the optimization problem expressed by~\equref{Equation:cmot_objective} is bi-level, which means the computation of target function depends on the solution to another nested optimization problem. By~\equref{Equation:ot_solution}, we can directly obtain the gradient~(or sub-gradient) of Sinkhorn distance $\mathrm{dis}(\mathbf{x}_i,\mathbf{x}_j)$ with respect to parameters in $\phi$, and make the whole {\name} easy to optimize. Refer to the supplement for more details.

\section{Experiments}
There are three parts of experiments in this section. In the first part, we test our method on a specially designed synthetic dataset to show the superiority of jointly considering semantic and temporal information. In the second part, we evaluate {\name} on three widely used benchmark datasets, {\em i.e.}, {\UCF}, {\HMDB}, and {\SMSM}. Last are further studies about {\name}. Experiment codes will be released after this paper being accepted.

\begin{table}[]
\centering
\begin{tabular}{@{}c|cc@{}}
\toprule
        & Content-Dominated & Ordering-Dominated \\
        & $5$-way $1$-shot  & $5$-way $1$-shot   \\ \midrule
TARN~\cite{TARN}    & 45.0 $\pm$ 0.4    & 33.2 $\pm$ 0.3     \\
OTAM~\cite{OTAM}    & 36.1 $\pm$ 0.4    & 46.3 $\pm$ 0.5     \\
{\name} & \bf 53.2 $\pm$ 0.6    & \bf 49.4 $\pm$ 0.4     \\ \bottomrule
\end{tabular}
\caption{Average accuracies ($\%$) with $95\%$ confidence intervals on tasks sampled from two different {\em meta-testing} sets.}
\label{Table:synthetic}
\end{table}

\subsection{Semantic-Temporal Trade-off}
\paragraph{Settings.} The most important advantage of {\name} is the ability to simultaneously consider semantic and temporal information. Moreover, we can achieve a trade-off between them by tuning hyper-parameter $\alpha$. To show this, we specially design a synthetic experiment in this part by reversing the frame sequences of videos. To be specific, we have a new video $\tilde{\mathbf{x}}$ by reversing video $\mathbf{x}$, and consider two different label assignments. \textbf{(1) Content-Dominated.} We assign the same label to $\tilde{\mathbf{x}}$ as $\mathbf{x}$. Some actions are not sensitive to frame ordering. For example, when reversing a video showing ``talk'', we still get a ``talk''. \textbf{(2) Ordering-Dominated.} We assign a new label to $\tilde{\mathbf{x}}$. Some actions have totally different meanings after being reversed. Intuitively, in the first setting, matching-based methods like OTAM won't work well. OTAM tries to find a strictly progressive matching between segment sequences, but now there exist two inverse videos with a same label, making it impossible for OTAM to pull them close in the metric space. In the second setting, aggregation-based methods will fail. Since typical aggregation-based methods generate a video-level representation by pooling and neglect the temporal information, they can't differ two videos with same frames but different orderings. However, our proposed {\name} can jointly consider semantic contents and temporal orderings, and it can achieve good performances in both settings.
\paragraph{Datasets.} We conduct this synthetic experiment on {\HMDB}. Firstly, we split {\HMDB} into {\em meta-training}, {\em meta-validation}, and {\em meta-testing} set. They contain $31$, $10$, and $10$ classes respectively. After that, we reverse videos in {\em meta-validation} set and {\em meta-testing} set to obtain new instances and assign labels to them. In \textbf{Content-Dominated} setting, a reversed video has a same label as its raw version, and the total number of classes in {\em meta-validation} set and {\em meta-testing} set keeps unchanged. In \textbf{Ordering-Dominated} setting, we assign new labels to reversed videos. Thus, the total number of classes in both {\em meta-validation} set and {\em meta-testing} set comes to $20$.
\paragraph{Results.} We implement an aggregation-based method TARN~\cite{TARN} and a matching-based method OTAM~\cite{OTAM}. We train these two models and our proposed {\name} on the {\em meta-training} set of {\HMDB}, and test them on different {\em meta-testing} set. We show the testing accuracy on two different {\em meta-testing} set in Table~\ref{Table:synthetic}. As expected, aggregation-based method fails the Ordering-Dominated setting while matching-based method fails the Content-Dominated setting. Our proposed method works well in both two settings, verifying the effectiveness of semantic-temporal trade-off.

\begin{table*}[]
\centering
\begin{tabular}{@{}c|c|cc|cc|cc@{}}
\toprule
\multirow{2}{*}{Method} & \multirow{2}{*}{Type}        & \multicolumn{2}{c|}{{\UCF}}         & \multicolumn{2}{c|}{{\HMDB}}        & \multicolumn{2}{c}{{\SMSM}}         \\
                        &                              & $5$-way $1$-shot & $5$-way $5$-shot & $5$-way $1$-shot & $5$-way $5$-shot & $5$-way $1$-shot & $5$-way $5$-shot \\ \midrule
TARN                    & \multirow{7}{*}{Aggregation} & \red{84.2 $\pm$ 0.4}   & \red{89.3 $\pm$ 0.6}   & \red{57.0 $\pm$ 0.2}   & \red{73.3 $\pm$ 0.3}   & \red{33.2 $\pm$ 0.2}   & \red{44.7 $\pm$ 0.4}   \\
TAEN                    &                              & \red{88.3 $\pm$ 0.4}   & \red{89.5 $\pm$ 0.4}   & \red{63.9 $\pm$ 0.4}   & \red{78.8 $\pm$ 0.3}   & \red{37.1 $\pm$ 0.4}   & \red{47.3 $\pm$ 0.4}   \\
ARN                     &                              & 62.1             & 84.8             & 44.6             & 59.1             & -   & -   \\
ITA                     &                              & 88.7             & \bf 96.7             & 63.4             & 79.6             & -   & -   \\
TAV                     &                              & 68.3             & 88.1             & 36.0             & 53.1             & -   & -   \\
AMeFu-Net               &                              & 85.1             & 95.5             & 60.2             & 75.5             & -   & -   \\
FAN                     &                              & 71.8             & 86.5             & 50.2             & 67.6             & -   & -   \\ \midrule
OTAM                    & Matching                     & \red{87.2 $\pm$ 0.3}   & \red{93.3 $\pm$ 0.5}   & \red{64.6 $\pm$ 0.5}   & \red{77.0 $\pm$ 0.3}   & 42.8             & 52.3             \\ \midrule
{\name}                 & Compromised                  & \bf \red{90.4 $\pm$ 0.4}   & \red{95.7 $\pm$ 0.3}   & \bf \red{66.9 $\pm$ 0.5}   & \bf \red{81.5 $\pm$ 0.4}   & \bf \red{46.8 $\pm$ 0.5}   & \bf \red{55.9 $\pm$ 0.4}   \\ \bottomrule
\end{tabular}
\caption{Average accuracies ($\%$) with $95\%$ confidence intervals on tasks sampled from {\em meta-testing} set of {\UCF}, {\HMDB}, and {\SMSM}. Best results are in bold. \red{\textbf{Red}} values are results from re-implementation. \textbf{Black} values are results cited from papers. For methods that use 3DCNN backbone, we re-implement them with pre-trained C3D. For methods that use CNN backbone, we re-implement them with ResNet-50~\cite{ResNet} pre-trained on ImageNet~\cite{ImageNet}.}
\label{Table:benchmark_results}
\end{table*}

\subsection{Benchmark Evaluation}
\begin{figure}[!t]
	\centering
	\includegraphics[width=\columnwidth]{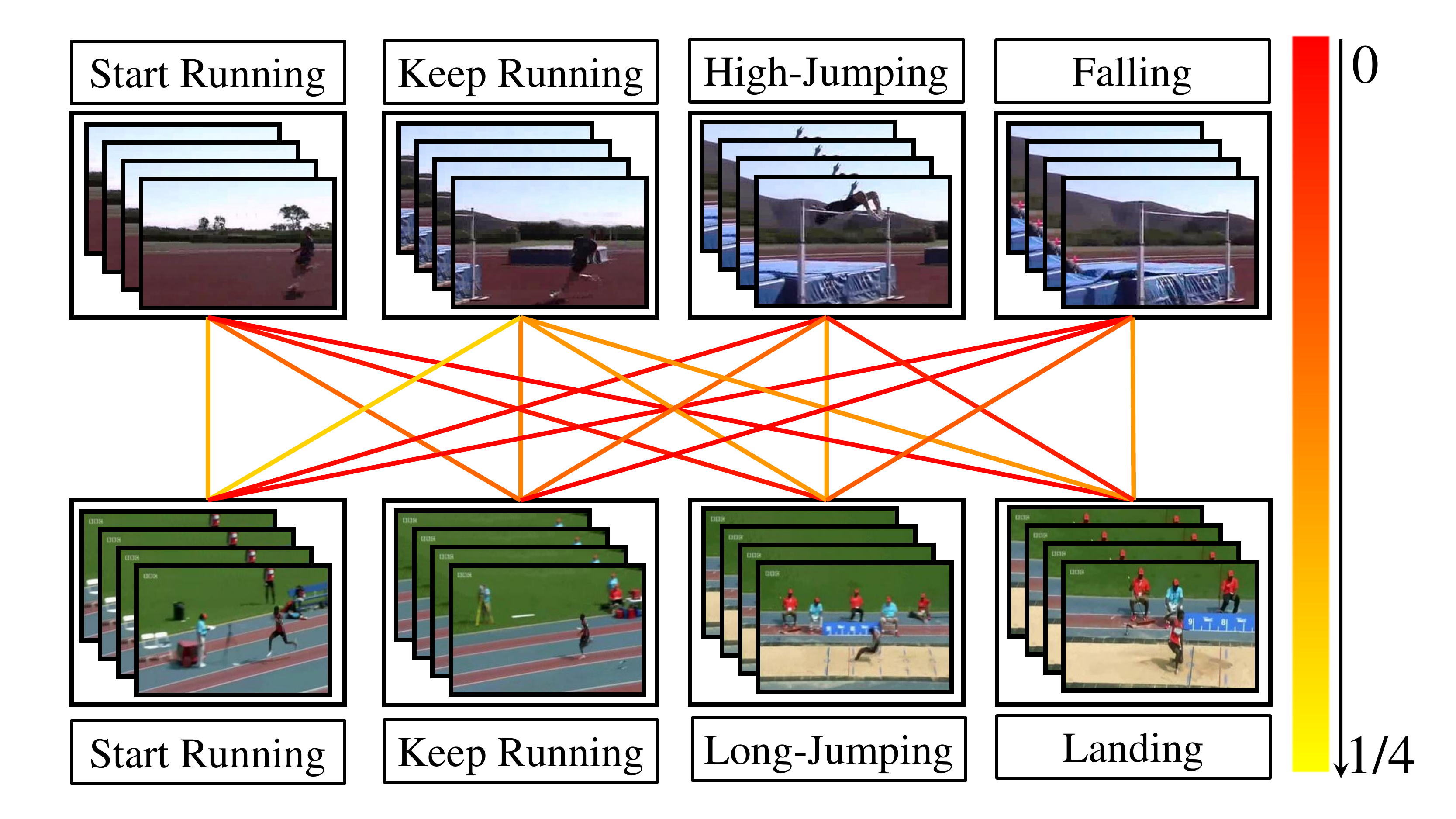}
	\caption{Visualization of transportation matrix between two videos sampled from {\UCF}. Each video is represented by $4$ segments. The color of edge represents the corresponding value in transportation matrix.}
	\label{Figure:visualization}
\end{figure}

\paragraph{Datasets.} {\UCF}~\cite{UCF-101}, {\HMDB}~\cite{HMDB-51}, and {\SMSM}~\cite{SMSM-V2} are three widely used datasets for action recognition. In our experiments, we re-sample and split them into {\em meta-training} set, {\em meta-validation} set, and {\em meta-testing} set to fit few-shot learning setting following~\cite{ARN,OTAM}. Some statistics about these datasets are shown in~\tabref{Table:dataset}. More information can be found in the supplement.
\begin{table}[!t]
\centering
\begin{tabular*}{\columnwidth}{@{}p{4cm}<{\centering}|p{1cm}<{\centering}p{1cm}<{\centering}p{1cm}<{\centering}@{}}
\toprule
                                   & UCF & HMDB & SMSM           \\ \midrule
\# classes                          & 101    & 51 & 100         \\
\# videos                           & 13320  & 6849 & 71718    \\
\# {\em meta-training} classes  & 70     & 31    & 64                \\
\# {\em meta-validation} classes & 10     & 10     & 12                 \\
\# {\em meta-testing} classes   & 21     & 10     & 24               \\ \bottomrule
\end{tabular*}
\caption{Statistics about three benchmark datasets.}
\label{Table:dataset}
\end{table}
\paragraph{Implementation Details.} As in many previous papers~\cite{ProtoGAN,ARN,TAEN,TARN}, we use C3D~\cite{C3D} pre-trained on Sports-1M~\cite{Sports-1M} as the embedding network for segment representations. For each video, we uniformly sample $M=4$ segments, each of which contains $16$ consecutive frames. Since we have no prior knowledge on these videos, the target distribution $\bm{\mu}$ in~\equref{Equation:cmot_objective} is set to an $4$-dimensional uniform distribution for all videos. In {\em meta-training} phase, we randomly sample $20000$ episodes from {\em meta-training} set. In {\em meta-testing} phase, we sample $1000$ episodes from {\em meta-testing} set, and report the average test accuracies with $95\%$ confidence intervals. More implementation details can be found in the supplementary material.
\paragraph{Results.} We compare our proposed {\name} to other few-shot action recognition methods. These methods can be classified into two categories, {\em i.e.}, aggregation-based methods and matching-based methods. We summarize experiment results in~\tabref{Table:benchmark_results}. {\name} achieves state-of-the-art performance in most cases. ITA~\cite{ITA} achieves best result on $5$-way $5$-shot {\UCF}, and this method utilizes an implicit segment alignment and considers temporal information to some extent, verifying that it is beneficial to incorporate sequence matching in few-shot action recognition. Different from {\UCF} and {\HMDB}, {\SMSM} dataset contains actions that are sensitive to temporal ordering, {\em e.g.}, putting something on a surface, moving something from left to right. Thus, matching-based method~(OTAM) achieves the highest performance among all the comparison methods. Our proposed {\name} simultaneously considers video contents and temporal information and outperforms OTAM.
\paragraph{Visualization of Transportation.} A key component in {\name} is optimal transport based on fused cost matrix $\mathbf{C}$, and it is necessary to check whether this module can output a reasonable matching based on both semantic and temporal information. \figref{Figure:visualization} shows the transportation matrix between two videos sampled from {\UCF}. We can see that segments with similar semantics are matched~(first and second segments in two videos), and segments at same position are matched. This means the transportation plan generated by {\name} jointly considers video contents and orderings.

\subsection{Further Analyses}
\paragraph{Ablation Study.} In this part, we evaluate the effectiveness of each module in {\name}. We sample tasks from {\HMDB} to evaluate several variants of {\name}. We summarize experiment results in~\tabref{Table:ablation}. By removing optimal transport module, our method degenerates to a simple metric-based meta-learning algorithm, which classifies the {\em query} instance to the class label of its nearest class center. Equipped with positional cost matrix and optimal transport, our model outperforms the baseline models by a noticeable margin.
\begin{table}[]
\centering
\small
\begin{tabular}{c|cccc}
\toprule
Model        & OT       & PO       & $5$-way $1$-shot & $5$-way $5$-shot \\ \midrule
0 & $\times$ & $\times$ & 63.4 $\pm$ 0.3   & 79.4 $\pm$ 0.3   \\
1  & $\checkmark$  & $\times$ & 64.9 $\pm$ 0.2   & 79.2 $\pm$ 0.1   \\
C3D $+$ LSTM & $\times$ & $\times$ & 51.8 $\pm$ 0.4   & 70.0 $\pm$ 0.3   \\
C3D $+$ DTW & $\times$ & $\times$ & 62.6 $\pm$ 0.4 & 77.3 $\pm$ 0.2 \\ \midrule
{\name}      & $\checkmark$  & $\checkmark$  & \bf 66.9 $\pm$ 0.5   & \bf 81.5 $\pm$ 0.4   \\ \bottomrule
\end{tabular}
\caption{Average test accuracies ($\%$) with $95\%$ confidence intervals of several variants on {\HMDB}. OT stands for optimal transport. PO means positional cost matrix. C3D $+$ LSTM means using an LSTM to capture temporal relations in segment features generated by C3D. C3D $+$ DTW means using DTW to align segment sequences, which is similar to OTAM (different embeddings).}
\label{Table:ablation}
\end{table}

\paragraph{Hyper-Parameter $\lambda$.} In {\name}, we utilize Sinkhorn distance~\equref{Equation:ot_sinkhorn} to measure the dissimilarity between two videos. Here $\lambda$ is a smoothing factor, and the larger it is, the closer Sinkhorn distance is to raw OT distance. Following~\cite{OT2}, we heuristically set $\lambda$ as $\{5,7,9\}\times 1/\mathrm{med}(\mathbf{C})$. Here $\mathrm{med}(\cdot)$ is the medium value operator and $\mathbf{C}$ is defined as~\equref{Equation:compromised_distance_ot}. Results are shown in~\tabref{Table:lambda}. We can see that the value of $\lambda$ has little influence on model accuracy.
\begin{table}[]
\centering
\small
\begin{tabular}{c|cc|cc}
\toprule
\multirow{2}{*}{$\lambda$} & \multicolumn{2}{c|}{{\UCF}}         & \multicolumn{2}{c}{{\HMDB}}        \\
                           & $1$-shot & $5$-shot & $1$-shot & $5$-shot \\ \midrule
$5\times\mathrm{med}$      & 89.0$\pm$0.3   & \bf 96.3$\pm$0.4   & 65.2$\pm$0.2   & 80.0$\pm$0.2   \\
\blue{$7\times\mathrm{med}$}      & \bf\blue{90.4$\pm$0.4}   & \blue{95.7$\pm$0.3}   & \bf\blue{66.9$\pm$0.5}   & \bf\blue{81.5$\pm$0.4}   \\
$9\times\mathrm{med}$      & 89.4$\pm$0.3   & 95.1$\pm$0.5   & 66.2$\pm$0.3   & 80.8$\pm$0.3   \\ \bottomrule
\end{tabular}
\caption{Average test accuracies ($\%$) with $95\%$ confidence intervals for different $\lambda$ values. Best results are in bold. \blue{Blue}: we choose \blue{$\lambda=7\times\mathrm{med}(\tilde{\mathbf{C}})$} in our experiments where $\mathrm{med}(\cdot)$ is medium value operator.}
\label{Table:lambda}
\end{table}

\begin{figure}[!t]
	\centering
	\includegraphics[width=\columnwidth]{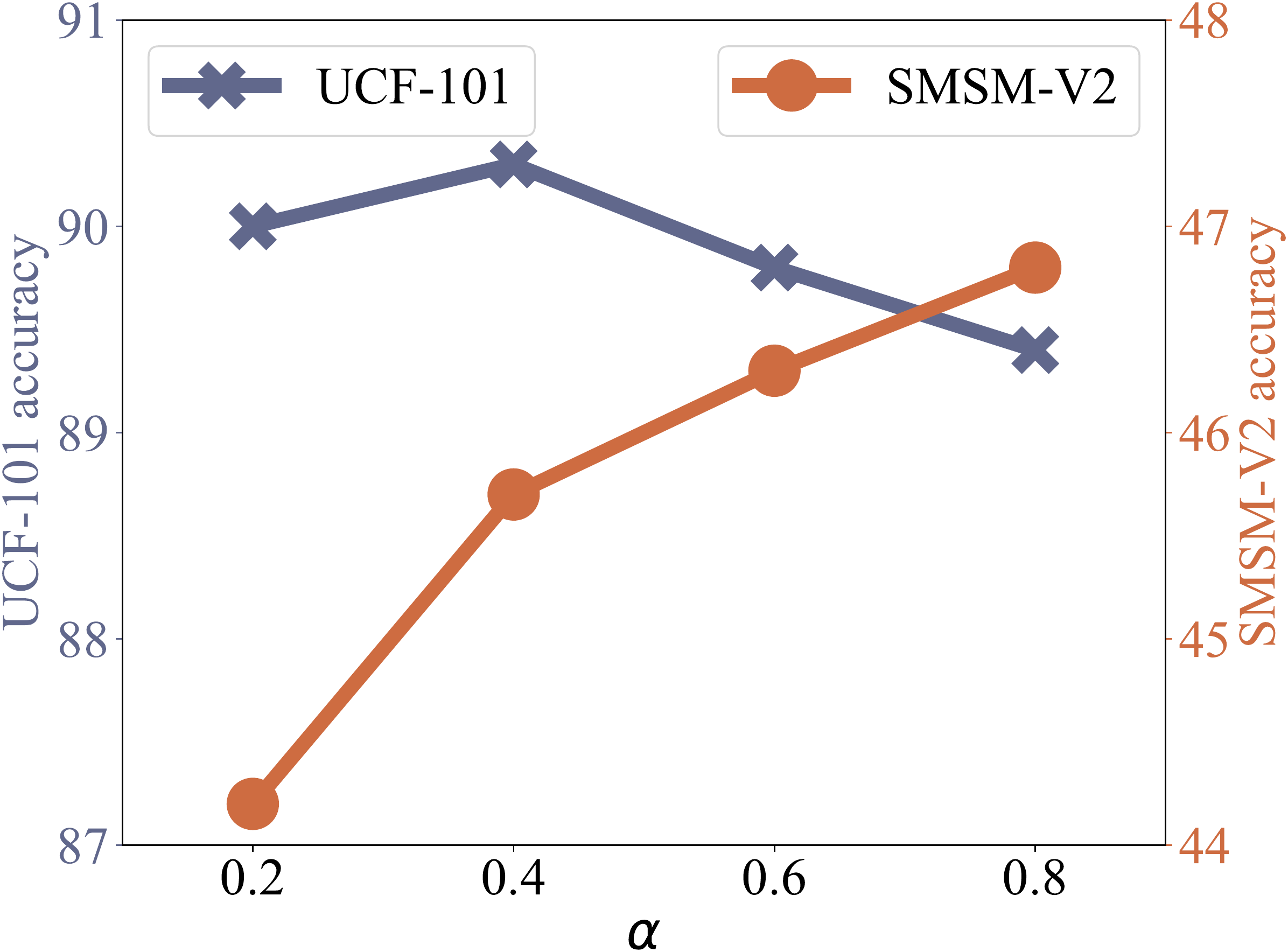}
	\caption{Average test accuracies ($\%$) on $5$-way $1$-shot tasks sampled from {\UCF} and {\SMSM}. We try different $\alpha$ values. Hyper-parameter $\alpha$ has relatively small influence on {\UCF} since it is a coarse-grained dataset. However, for {\SMSM}, we should pay more attention to long-term temporal ordering to obtain a higher accuracy.}
	\label{Figure:alpha}
\end{figure}

\paragraph{Hyper-parameter $\alpha$.} $\alpha$ balances the importance of semantic contents and temporal orderings. Intuitively, for tasks that are more sensitive to temporal orderings, we should set $\alpha$ to a larger value. As indicated in the previous subsection, {\SMSM} dataset contains fine-grained actions that are not easy to differ without considering long-term orderings. Thus, we check the influence of $\alpha$ on {\UCF} and {\SMSM}. Results are shown in Figure~\ref{Figure:alpha}.

\section{Conclusion}
In this paper, we propose \textbf{C}ompromised \textbf{M}etric via \textbf{O}ptimal \textbf{T}ransport~({\name}), a new framework that jointly considers video contents and temporal relations in few-shot action recognition. Different from existing methods that either focus on semantic information by aggregation or emphasize temporal information by sequence alignment, we simultaneously take two aspects into account within a single OT problem. {\name} defines two videos' distance as the transportation cost between their segment sequences based on a fused cost matrix, and then predicts the label using a metric-based meta-learning method. On several benchmark datasets, {\name} achieves state-of-the-art results.

\section*{Appendix A: Review of {\name}}
{\name} is designed for few-shot action recognition problems. It jointly considers semantic and temporal information in videos and unites them in an Optimal Transport~(OT) framework. In this section, we will first give a concise review of {\name}, and then discuss several details.

\subsection*{Algorithm Review}
{\name} belongs to metric-based meta-learning methods~\cite{Siamese, MatchNet, ProtoNet,MetaOptNet,FEAT}. These algorithms often meta-learn an embedding function from {\em meta-training} set, and then reuse the learned representation on {\em meta-testing} set. Obviously, a key point in applying these methods to few-shot action recognition is the design of distance metric function. In few-shot action recognition, it is more difficult to measure the distance between videos than images owing to the complex temporal structures.

\begin{algorithm}[t]
	\begin{algorithmic}{
			\REQUIRE Cost matrix $\mathbf{C}$, target distribution $\bm{\mu}$ and $\bm{\nu}$, smoothing parameter $\lambda$.\\
			\STATE Compute $G=\exp(-\lambda\mathbf{C})$.
			\STATE Initialize $\mathbf{u}$ and $\mathbf{v}$.
			\REPEAT
				\STATE Compute $\mathbf{u}=\bm{\mu}\oslash(\mathbf{G}\mathbf{v})$.
				\STATE Compute $\mathbf{v}=\bm{\nu}\oslash(\mathbf{G}^\top\mathbf{u})$.
			\UNTIL{Convergence}
			\STATE Compute $\mathbf{T}^\star=\text{diag}(\mathbf{u})\mathbf{G}\text{diag}(\mathbf{v})$.
		}
	\end{algorithmic}
	\caption{Algorithm flow for solving Sinkhorn problem.}
	\label{Algorithm:sinkhorn}
\end{algorithm}

Existing few-shot action recognition algorithms can be classified into two categories, {\em i.e.}, \textbf{aggregation-based} methods and \textbf{matching-based} methods. Both of these two kinds of methods first sample frames or segments from a video and get their content representations, but aggregation-based methods generate a video-level representation for distance calculation by pooling~\cite{MB,ProtoGAN,3DFSV,PAL,TAEN,ITA}, adaptive fusion~\cite{AMeFu,TAV}, dynamic images~\cite{FAN}, or attention~\cite{TARN,ARN,PASTN} while matching-based methods explicitly align two sequences and define distance between videos as their matching cost by DTW~\cite{OTAM} or other matching strategies. Aggregation-based methods tend to focus on semantic information while matching-based methods impose a strict regulation on video orderings.

In {\name}, we simultaneously consider semantic and temporal information in videos. We first sample segments from videos and get their content representations by 3DCNN~\cite{C3D,I3D}, and then compute the distance matrix between two segment sequences to form a semantic cost matrix. After that, we define the positional cost matrix as the relative distance between two segments' positions. By adding the positional cost matrix to segment cost matrix, we can obtain a fused cost matrix, the optimal transportation plan based on which jointly considers video contents and segment orderings.

\subsection*{Sinkhorn Algorithm}
An important step in computing video distances with {\name} is solving the OT problem in~\equref{Equation:compromised_distance_ot}.
\begin{equation}
\begin{aligned}
&\mathrm{dis}(\mathbf{x}_1,\mathbf{x}_2)&&=\min_{\mathbf{T}\in\Pi(\bm{\mu}_{1},\bm{\mu}_{2})}\langle\mathbf{T},\mathbf{C}\rangle-\frac{1}{\lambda}\mathcal{H}(\mathbf{T})\\
&\mathrm{s.t.}&&\mathbf{C}=\mathbf{C}^{\mathrm{SE}}+\alpha\mathbf{C}^{\mathrm{PO}}
\end{aligned}
\label{Equation:compromised_distance_ot}
\end{equation}
Here $\mathbf{C}^{\mathrm{SE}}$ is the semantic cost matrix and $\mathbf{C}^{\mathrm{PO}}$ is the positional cost matrix. $\bm{\mu}_1$ and $\bm{\mu}_2$ are target distributions for two videos, and we empirically set them as uniform $M$-dimensional distributions. OT problem is not easy to solve, but by adding an entropy term $\frac{1}{\lambda}\mathcal{H}(\mathbf{T})$, we make the objective function smoother and have efficient numerical algorithms~\cite{OT2} as shown in~\algref{Algorithm:sinkhorn}.

After executing~\algref{Algorithm:sinkhorn}, we obtain a `closed form' of optimal transportation plan $\mathbf{T}^\star=\text{diag}(\bm{\mu}_1)\mathbf{G}\text{diag}(\bm{\mu}_2)$, and can differentiate the transportation cost w.r.t. cost matrix $\mathbf{C}$. The learnable parameters contained in 3DCNN can be updated by sub-gradient descent.

\begin{table*}[]
\centering
\begin{tabular}{@{}c|cc|cc@{}}
\toprule
\multirow{2}{*}{Positional Cost Matrix} & \multicolumn{2}{c|}{{\UCF}}         & \multicolumn{2}{c}{{\HMDB}}        \\
                                        & $5$-way $1$-shot & $5$-way $5$-shot & $5$-way $1$-shot & $5$-way $5$-shot \\ \midrule
Induced by Uniform PE                   & 89.1 $\pm$ 0.3   & 93.9 $\pm$ 0.2   & 65.2 $\pm$ 0.4   & 79.3 $\pm$ 0.3   \\
Induced by Sinusoid PE                  & 87.1 $\pm$ 0.3   & 91.2 $\pm$ 0.3   & 64.1 $\pm$ 0.3   & 78.1 $\pm$ 0.5   \\
Our Proposal                            & \bf 90.4 $\pm$ 0.4   & \bf 95.7 $\pm$ 0.3   & \bf 66.9 $\pm$ 0.5   & \bf 81.5 $\pm$ 0.4   \\ \bottomrule
\end{tabular}
\caption{Average accuracies ($\%$) with $95\%$ confidence intervals on tasks sampled from {\em meta-testing} set of {\UCF} and {\HMDB}. Best results are in bold. Our proposed positional cost matrix achieves best results.}
\label{Table:positional_encoding}
\end{table*}

\subsection*{Positional Cost Matrix}
In order to capture long-term temporal relations in videos, {\name} additionally penalize the semantic cost matrix with relative distances between segments, as shown in~\equref{Equation:positional_cost_matrix}. This means the transportation cost between distant segments will be enlarged, encouraging OT algorithm to assign a lower value to the corresponding transportation plan.
\begin{equation}
\mathbf{C}^{\mathrm{PO}}_{pq}=\exp\left\{-\frac{1}{\sigma^2}\frac{1}{\left(\frac{p}{M}-\frac{q}{M}\right)^2+1}\right\},\forall p,q\in[M]
\label{Equation:positional_cost_matrix}
\end{equation}

\begin{figure}[!t]
	\centering
	\includegraphics[width=\columnwidth]{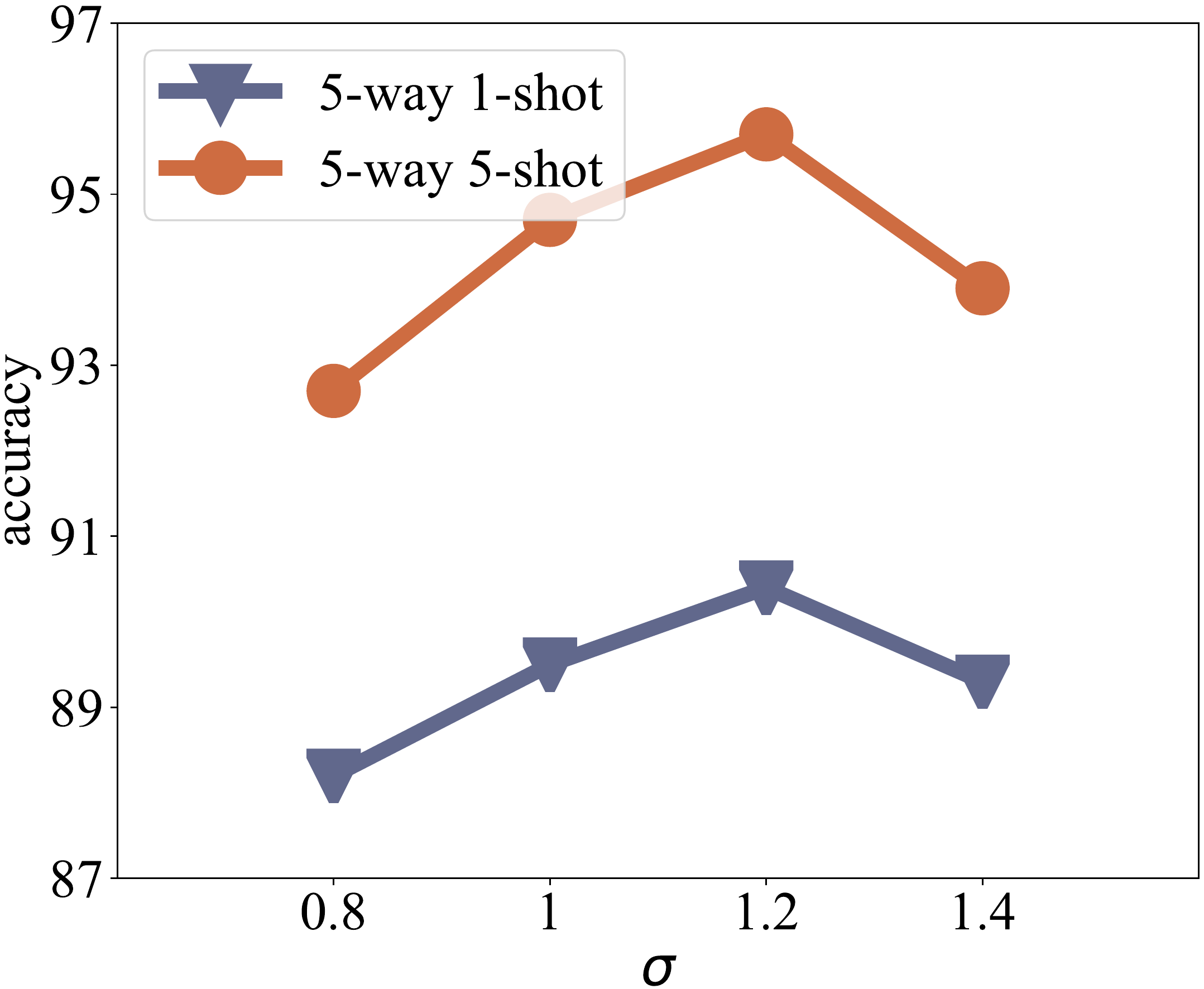}
	\caption{Average test accuracies ($\%$) on $5$-way $1$-shot tasks and $5$-way $5$-shot tasks sampled from {\UCF}. We try different $\sigma$ values.}
	\label{Figure:sigma}
\end{figure}

In~\equref{Equation:positional_cost_matrix}. $\sigma$ is a hyper-parameter that rescales the values in positional cost matrix. We check the influence of $\sigma$ on {\UCF} dataset. The results are shown in~\figref{Figure:sigma}. We can see that $\sigma$ has a non-neglectable influence on model performance. A larger $\sigma$ makes the distribution of values in $\mathbf{C}$ smoother, weakening the impact of temporal relations between segment sequences. For tasks that are sensitive to video orderings, we are supposed to set a smaller $\sigma$.

Positional cost matrix can be defined as other formulations. For example, we can borrow the idea of positional encoding~\cite{Transformer}, and defines the transportation cost between two segments as the difference between their positional encodings. Assuming that a video is composed of $M$ segments and the dimension of positional encoding is $D$, there are two commonly used positional encodings:
\begin{itemize}
\item Uniform Positional Encodings. $\mathbf{P}_{md}=\frac{m}{M},\forall m\in[M],\forall d\in[D]$.
\item Sinusoid Positional Encodings. $\mathbf{P}_{md}=\left\{\begin{matrix}
\sin\left(\frac{m}{10000^{d/D}}\right), & d\text{ is even}\\ 
\cos\left(\frac{m}{10000^{(d-1)/D}}\right), & d \text{ is odd} 
\end{matrix}\right.,\forall m\in[M],\forall d\in[D]$.
\end{itemize}
Positional cost matrix can be computed as $\mathbf{C}^{\mathrm{PO}}_{pq}=\|\mathbf{P}_{p}-\mathbf{P}_{q}\|_2$. For uniform positional encodings, we have $\mathbf{C}^{\mathrm{PO}}=\frac{\sqrt{D}}{M}|p-q|$, which is monotonically increasing in the relative temporal distance $|p-q|$. Thus, the positional cost matrix induced by uniform positional encodings have similar properties as our proposed one. However, sinusoid positional encodings proposed in~\cite{Transformer} don't have this property, and is not very suitable for modelling long-term temporal relations in our case. \tabref{Table:positional_encoding} gives the results of an ablation study on positional cost matrix. As expected, uniform positional encoding outperforms sinusoid positional encoding.

\subsection*{Segment Sampling Strategy}
All the existing few-shot action recognition methods first sample segments from the whole video, and sampling strategy is a vital factor to model performance. However, there lacks a systematic research of sampling strategy. After summarizing existing papers~\cite{MB,ProtoGAN,3DFSV,PAL,TAEN,ITA,AMeFu,TAV,FAN,OTAM,TARN,ARN,PASTN}, we divide sampling strategies into two categories:
\begin{itemize}
\item Cut the video into $M$ clips and sample one single frame from each clip. In this case, a segment is composed of a single image.
\item Cut the video into $M$ clips and treat each clip as a segment. In this case, a segment contains many consecutive images.
\end{itemize}
The second strategy includes redundant information since consecutive images are similar to each other while the first strategy may ignore some important frames. In {\name}, after cutting the video into $M$ clips, we randomly sample $16$ frames from each clip to form a segment, balancing intra-segment and inter-segment relations. \tabref{Table:sampling_strategy} is an ablation study about sampling strategies.

\begin{table}[]
\centering
\begin{tabular}{@{}c|cc@{}}
\toprule
\multirow{2}{*}{\begin{tabular}[c]{@{}c@{}}Sampling\\ Strategy\end{tabular}} & \multicolumn{2}{c}{{\SMSM}}         \\
                                                                             & $5$-way $1$-shot & $5$-way $5$-shot \\ \midrule
Single Frame                                                                 & 40.3 $\pm$ 0.4   & 47.9 $\pm$ 0.4   \\
Whole Clip                                                                   & 44.2 $\pm$ 0.3   & 51.7 $\pm$ 0.5   \\
Our Proposal                                                                 & \bf 46.8 $\pm$ 0.5   & \bf 55.9 $\pm$ 0.4   \\ \bottomrule
\end{tabular}
\caption{Average test accuracies ($\%$) on $5$-way $1$-shot tasks and $5$-way $5$-shot tasks sampled from {\SMSM}. We try three different sampling strategies.}
\label{Table:sampling_strategy}
\end{table}

\section*{Appendix B: Experiment Details}\label{Section:experiment_details}
In this part, we give more information about experiments, such as dataset split and implementation details.
\subsection*{Datasets}
In this paper, we use three action recognition benchmark datasets, {\em i.e.}, {\UCF}~\cite{UCF-101}, {\HMDB}~\cite{HMDB-51}, and {\SMSM}~\cite{SMSM-V2} to evaluate {\name} and comparison methods. We split each dataset into {\em meta-training} set, {\em meta-validation} set, and {\em meta-testing} set to fit few-shot learning setting. {\UCF} contains 101 classes while {\HMDB} contains 51 classes. These two datasets contain coarse-grained actions that are more sensitive to video contents. {\SMSM} contains 100 classes, and is composed of fine-grained actions like `moving something from left to right'. \tabref{Table:UCF_split},~\tabref{Table:HMDB_split}, and~\tabref{Table:SMSM_split} show detailed split of these three datasets.
\begin{table*}[]
\centering
\begin{tabular}{@{}c|c@{}}
\toprule
split                 & classes \\ \midrule
{\em meta-training}   & \tabincell{c}{IceDancing, CricketShot, FieldHockeyPenalty, RockClimbingIndoor, CricketBowling, BaseballPitch,\\CuttingInKitchen, PlayingDhol, MilitaryParade, BrushingTeeth, PoleVault, VolleyballSpiking,\\Hammering, 
Surfing, HeadMassage, Knitting, SkyDiving, HammerThrow,\\TaiChi, PlayingTabla, JumpingJack, ApplyLipstick, PlayingCello, BoxingPunchingBag,\\Haircut, Drumming, BodyWeightSquats, ParallelBars, JavelinThrow, FrisbeeCatch,\\TennisSwing, BreastStroke, FloorGymnastics, RopeClimbing, TableTennisShot, Nunchucks,\\HorseRiding, GolfSwing, Skijet, HorseRace, Skiing, JugglingBalls,\\PommelHorse, Archery, BalanceBeam, SoccerPenalty, TrampolineJumping, PlayingDaf,\\BandMarching, Fencing, PushUps, PlayingSitar, BoxingSpeedBag, JumpRope,\\PlayingGuitar, SalsaSpin, Mixing, CleanAndJerk, Typing, Diving,\\WalkingWithDog, WritingOnBoard, HighJump, PlayingViolin, Rowing, PlayingFlute,\\Rafting, BabyCrawling, SoccerJuggling, BenchPress}       \\ \midrule
{\em meta-validation} & \tabincell{c}{Biking, Billiards, Basketball, ShavingBeard, PizzaTossing, HandstandWalking,\\BlowDryHair, SkateBoarding, PullUps, HandstandPushups}        \\ \midrule
{\em meta-testing}    & \tabincell{c}{FrontCrawl, Shotput, LongJump, BlowingCandles, BasketballDunk, ThrowDiscus,\\UnevenBars, Swing, Bowling, HulaHoop, Lunges, StillRings,\\WallPushups, ApplyEyeMakeup, PlayingPiano, CliffDiving, SumoWrestling, MoppingFloor,\\YoYo, Punch, Kayaking}        \\ \bottomrule
\end{tabular}
\caption{Dataset split of {\UCF}.}
\label{Table:UCF_split}
\end{table*}

\begin{table*}[]
\centering
\begin{tabular}{@{}c|c@{}}
\toprule
split                 & classes \\ \midrule
{\em meta-training}   & \tabincell{c}{shoot-bow, jump, smoke, smile, fall-floor, dribble,\\turn, walk, shoot-ball, sit, sword, kiss,\\throw, cartwheel, pushup, sword-exercise, run, pour,\\eat, draw-sword, catch, kick-ball, climb-stairs, pullup,\\hit, stand, handstand, swing-baseball, push, talk,\\clap}       \\ \midrule
{\em meta-validation} & \tabincell{c}{ride-bike, golf, flic-flac, kick, dive, chew,\\ride-horse, shoot-gun, climb, pick}        \\ \midrule
{\em meta-testing}    & \tabincell{c}{laugh, wave, shake-hands, punch, situp, brush-hair,\\drink, fencing, somersault, hug}        \\ \bottomrule
\end{tabular}
\caption{Dataset split of {\HMDB}.}
\label{Table:HMDB_split}
\end{table*}

\begin{table*}[]
\centering
\begin{tabular}{@{}c|c@{}}
\toprule
split                 & classes \\ \midrule
{\em meta-training}   & \tabincell{c}{Dropping sth behind sth, Poking a stack of sth so the stack collapses, Spilling sth next to sth, \\Failing to put sth into sth because sth does not fit, Pretending to throw sth, Taking sth out of sth, \\Moving away from sth with your camera, Pretending to take sth out of sth, Showing sth on top of sth, \\Lifting up one end of sth, then letting it drop down, Pretending to put sth on a surface, \\Pushing sth with sth, Pulling sth from right to left, Rolling sth on a flat surface, \\Moving sth away from the camera, Pretending to put sth underneath sth, Tearing sth into two pieces, \\Putting sth onto sth else that cannot support it so it falls down, Showing sth to the camera, \\Dropping sth next to sth, Lifting a surface with sth on it until it starts sliding down, \\Spinning sth so it continues spinning, Tilting sth with sth on it until it falls off, Plugging sth into sth, \\Pretending to be tearing sth that is not tearable, Pretending to sprinkle air onto sth, \\Lifting up one end of sth without letting it drop down, Turning the camera upwards while filming sth, \\Tilting sth with sth on it slightly so it doesn't fall down, Pretending to turn sth upside down, \\Lifting a surface with sth on it but not enough for it to slide down, Unfolding sth, \\Moving sth across a surface until it falls down, Poking a hole into some substance, \\Spinning sth that quickly stops spinning, Covering sth with sth, Putting sth next to sth, \\Putting sth and sth on the table, Throwing sth onto a surface, Pushing sth from left to right, \\Plugging sth into sth but pulling it right out as you remove your hand, \\Sth colliding with sth and both are being deflected, Laying sth on the table on its side, not upright, \\Poking a hole into sth soft, Pushing sth so that it slightly moves, Putting sth onto sth, \\Holding sth over sth, Letting sth roll along a flat surface, Folding sth, \\Trying to pour sth into sth, but missing so it spills next to it, Showing a photo of sth to the camera, \\Tipping sth with sth in it over, so sth in it falls out, Twisting (wringing) sth wet until water comes out, \\Taking sth from somewhere, Throwing sth against sth, Spilling sth behind sth, \\Approaching sth with your camera, Pretending to squeeze sth, Putting sth behind sth, Holding sth behind sth, \\Poking sth so that it falls over, Pretending to put sth behind sth, Pulling sth onto sth, \\Spreading sth onto sth}       \\ \midrule
{\em meta-validation} & \tabincell{c}{Putting sth, sth and sth on the table, Putting sth onto sth else that cannot support it so it falls down, \\Pulling two ends of sth so that it separates into two pieces, Pulling sth from right to left, \\Dropping sth into sth, Pouring sth onto sth, Pretending or failing to wipe sth off of sth, Folding sth, \\Spinning sth so it continues spinning, Pushing sth so that it falls off the table, \\Letting sth roll down a slanted surface, Showing sth to the camera}        \\ \midrule
{\em meta-testing}    & \tabincell{c}{Spreading sth onto sth, Pretending to throw sth, Pretending to close sth without actually closing it, \\Spinning sth that quickly stops spinning, Pushing sth off of sth, Throwing sth, Pushing sth onto sth, \\Taking one of many similar things on the table, Holding sth over sth, Putting sth upright on the table, \\Tilting sth with sth on it until it falls off, Throwing sth in the air and catching it, \\Showing that sth is inside sth, Pretending to open sth without actually opening it, \\Pretending to turn sth upside down, Pouring sth out of sth, Throwing sth in the air and letting it fall, \\Showing that sth is empty, Putting sth and sth on the table, Showing sth to the camera, \\Sprinkling sth onto sth, Opening sth, Tearing sth into two pieces, \\Moving sth across a surface until it falls down}        \\ \bottomrule
\end{tabular}
\caption{Dataset split of {\SMSM}.}
\label{Table:SMSM_split}
\end{table*}

\subsection*{Implementation Details}
In this section, we give more implementation details of {\name}. As in many previous works~\cite{ProtoGAN,ARN,TAEN,TARN}, we use C3D~\cite{C3D} pre-trained on Sports-1M~\cite{Sports-1M} as the embedding network for frame segment. For each video, we uniformly sample $M=4$ segments, each of which contains $16$ consecutive frames. The target distribution $\bm{\mu}$ is set to an $M$-dimensional uniform distribution for all videos. In {\em meta-training} phase, we randomly sample $20000$ episodes from {\em meta-training} set. In {\em meta-testing} phase, we sample $1000$ episodes from {\em meta-testing} set. For an $N$-way $K$-shot task, we sample $1$ video for each class to form the {\em query} set. We use SGD optimizer to train the model. The initial learning rate for pre-trained segment descriptor is set to $0.001$, and the initial learning rate for other parameters is set to $0.01$. Two learning rates are decreased by $0.2$ after every $2000$ episodes. As indicated in the paper, smoothing hyper-parameter $\lambda$ is set to $7\times\mathrm{med}(\mathbf{C})$ where $\mathrm{med}(\cdot)$ is the medium value operator and $\mathbf{C}$ is the fused cost matrix. Hyper-parameter $\alpha$ is set as $0.4$, $0.4$, and $0.8$ for {\UCF}, {\HMDB}, and {\SMSM} respectively. Hyper-parameter $\sigma$ is set as $1.2$ for three benchmark datasets.

{\small
\bibliographystyle{ieee_fullname}
\bibliography{references}
}

\end{document}